\documentclass[letterpaper, 10 pt, conference]{ieeeconf}  
\usepackage{graphicx} 
\usepackage{amssymb}
\usepackage[bookmarks=true]{hyperref}
\usepackage{amsmath}
\usepackage{url}
\usepackage[table,xcdraw]{xcolor}
\usepackage{booktabs}
\usepackage{xcolor}
\IEEEoverridecommandlockouts                              

\title{\LARGE \bf
SR3D: Unleashing Single-view 3D Reconstruction for Transparent and Specular Object Grasping
}

\author{
Mingxu Zhang$^{1*}$, 
Xiaoqi Li$^{2*}$, 
Jiahui Xu$^{1*}$, 
Kaichen Zhou$^{2}$, 
Hojin Bae$^{2}$, 
Yan Shen$^{2}$, 
\\Chuyan Xiong$^{3}$, 
Hao Dong$^{2}$
\vspace{0.1cm}\\
\textsuperscript{\rm 1} Beijing University of Posts and Telecommunications \\
\textsuperscript{\rm 2} Center on Frontiers of Computing Studies, Peking University \\
\textsuperscript{\rm 3}Institute of Computing Technology, Chinese Academy of Sciences
\thanks{*Equal contribution}
\thanks{$^{1}$Beijing University of Posts and Telecommunications}%
\thanks{$^{2}$Center on Frontiers of Computing Studies, Peking University}%
\thanks{$^{3}$Institute of Computing Technology, Chinese Academy of Sciences}
}

\begin{document}

\maketitle
\thispagestyle{empty}
\pagestyle{empty}

\begin{abstract}
Recent advancements in 3D robotic manipulation have improved grasping of everyday objects, but transparent and specular materials remain challenging due to depth sensing limitations. 
While several 3D reconstruction and depth completion approaches address these challenges, they suffer from setup complexity or limited observation information utilization. 
To address this, leveraging the power of single-view 3D object reconstruction approaches, we propose a training-free framework SR3D that enables robotic grasping of transparent and specular objects from a single-view observation. 
Specifically, given single-view RGB and depth images, SR3D first uses the external visual models to generate 3D reconstructed object mesh based on RGB image.
Then, the key idea is to determine the 3D object’s pose and scale to accurately localize the reconstructed object back into its original depth corrupted 3D scene.
Therefore, we propose \textbf{\textit{view matching}} and \textbf{\textit{keypoint matching}} mechanisms, which leverage both the 2D and 3D's  inherent semantic and geometric information in the observation to determine the object's 3D state within the scene, thereby reconstructing an accurate 3D depth map for effective grasp detection.
Experiments in both simulation and real-world show the reconstruction effectiveness of SR3D.
More demonstrations can be found at: \url{https://sites.google.com/view/sr3dtech/}
\end{abstract}

\section{INTRODUCTION}
In recent years, due to the 3D nature of interactive environments, 3D robotic manipulation has become crucial, and significant advancements~\cite{gervet2023act3d,goyal2023rvt,goyal2024rvt,shridhar2023perceiver,ze20243d,grotz2024peract2,li2024manipllm,liu2024robomamba,sundermeyer2021contact,jia2024lift3d} have been made in the field of robotic grasping and manipulation, particularly in the context of interacting with everyday objects.
However, transparent and specular objects such as glass and plastic continue to pose substantial challenges for robotic systems that depend on accurate depth sensing. 
Conventional depth sensors~\cite{keselman2017intel}, widely used in opaque object manipulation~\cite{fang2023anygrasp,sundermeyer2021contact,lu2023vl,mosbach2024grasp}, fail to reliably capture depth information for transparent materials.
In fact, studies show standard depth sensors like Intel RealSense achieves less than 55\% valid pixel coverage on clear glass surfaces~\cite{curto2022experimental}. 

A number of recent robotic manipulation methods have attempted to address this challenge~\cite{sajjan2020clear,fang2022transcg,dai2022domain,sajjan2020clear,wang2021graspness,breyer2021volumetric,fang2020graspnet}.
On one hand, some approaches~\cite{dai2023graspnerf} reconstruct the multi-view image to obtain 3D scene representation and show promising performance for transparent and specular object manipulation. However, they usually require 4–8 calibrated camera views for 3D reconstruction, leading to impractical requirements and increased setup complexity.
On the other hand, some approaches~\cite{shi2024asgrasp, fang2022transcg} rely on single-view or infrared (IR) image pairs and use depth completion or reconstruction techniques to recover incomplete depth maps. 
\begin{figure}
\includegraphics[width=0.5\textwidth]{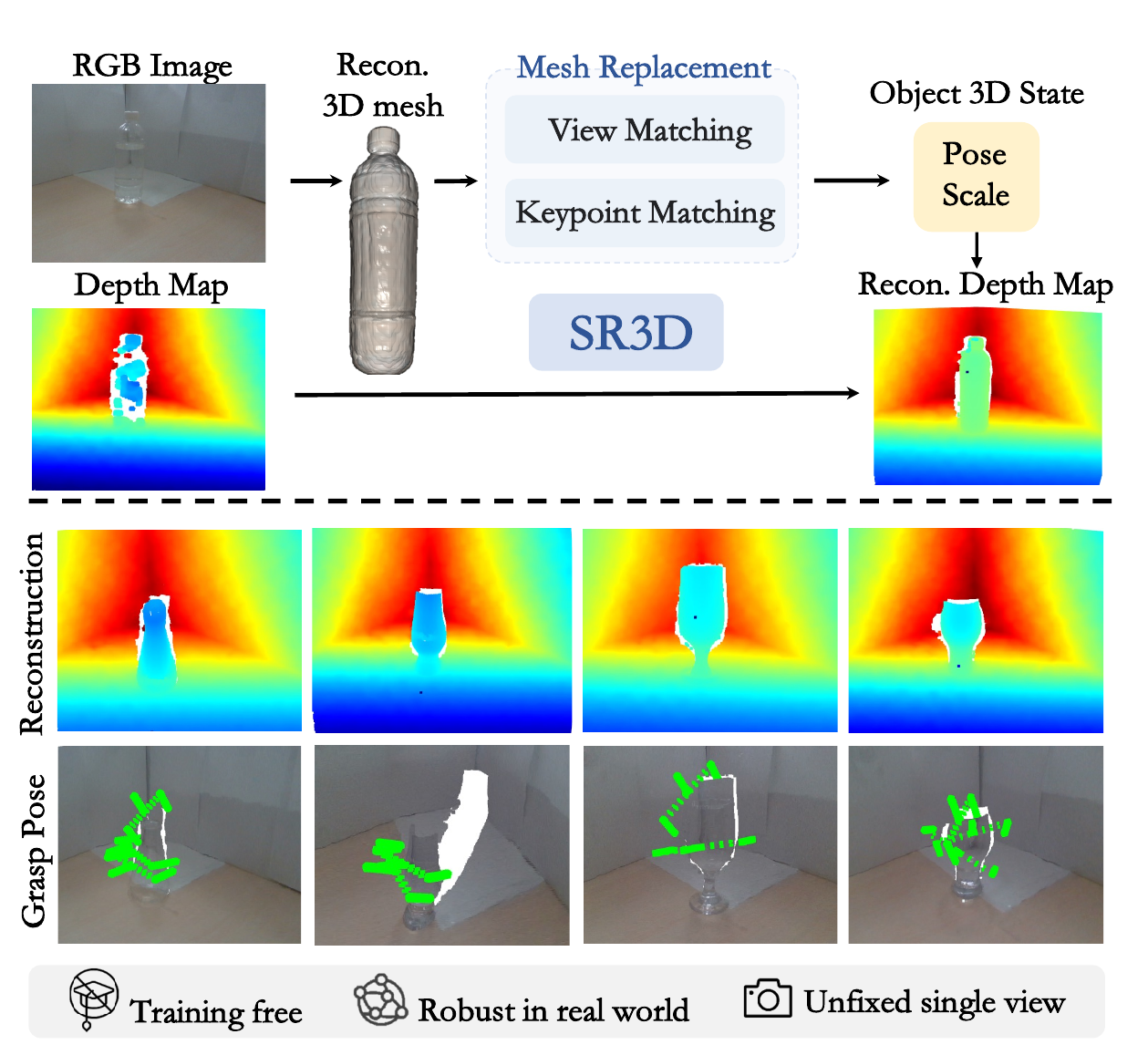}
    \caption{The top part illustrates the pipeline of SR3D, which takes single-view RGB-D observation to realize depth map reconstruction for transparent or specular objects. The bottom part demonstrates its effectiveness for robotic grasping on transparent objects. }
    \label{fig:teaser}
\end{figure}
However, these methods learn a direct mapping from 2D images to 3D reconstructions, lacking a deeper understanding that could be gained from precise depth data for opaque regions and the 2D semantic or geometric information extracted from transparent or specular objects.


Moreover, in the field of general single-image 3D reconstruction, object reconstruction methods~\cite{tang2023dreamgaussian,liu2023one,liu2024one,huang2024zeroshape,tochilkin2024triposr,zou2024triplane,he2023openlrm,huang2024zeroshape} have shown superior performance across a variety of single-camera view angles and object instances, offering efficient inference speeds, even on transparent and specular objects. 
This presents a promising approach for generating accurate 3D meshes for real-world robotic manipulation. 
However, these methods generally focus on geometry understanding and struggle to predict the required \textit{pose} and \textit{scale} needed to integrate the reconstructed object into the original world-coordinate 3D scene for effective robotic manipulation.

Inspired by the above, as shown in Fig.~\ref{fig:teaser}, we propose a training-free framework that harnesses the generalization ability of single-view object reconstruction techniques to enable reliable manipulation of transparent and specular objects from a single-view observation.
Specifically, given the RGB and depth observations from a single view, we leverage external visual models (e.g., Grounding-SAM~\cite{liu2024grounding}) to segment transparent or specular objects in the 2D image and employ single-view object reconstruction (e.g., TripoSR~\cite{tochilkin2024triposr}) methods to generate a 3D object mesh. 
However, the object's pose and scale, which are needed to align it with the original 3D scene, remain unknown.
We observe that in RGB-D observations, only the transparent or specular regions suffer from corrupted depth data, while the remaining opaque regions (e.g., the supporting table) retain accurate depth information.
Meanwhile, the 2D semantic and geometric information of objects present in the image still provides valuable cues that can be effectively leveraged.

Therefore, our goal is to utilize the reliable depth data, along with the crucial semantic and geometry cues from 2D image, to accurately determine the 3D object's state in the scene, including its \textit{pose} and \textit{scale}. 
To achieve this, we introduce two key mechanisms: \textbf{view matching} and \textbf{keypoint matching}.
In \textbf{view matching}, we virtually project the reconstructed 3D object onto multiple viewpoints, compute the 2D structural and geometric similarity between each rendered object image and the captured 2D image, and select the viewpoint with the highest similarity score as the object’s orientation in the 3D scene.
In \textbf{keypoint matching}, our goal is to determine the object's position and scale. We first automatically detect 2D keypoints on critical structural regions, such as the contact edges between the object and the supporting opaque table.
Given the 3D positional consistency of the contact edge lines between the object and the supporting surface (e.g., a table), we align the corresponding 3D keypoints to determine the object's position and scale in the 3D scene.
These allow us to integrate the 3D object mesh into the captured 3D scene, after which we apply a grasp detection model~\cite{fang2023anygrasp} to ensure an effective robotic grasping.

Our experiments, conducted in both simulation and real-world, demonstrate that the proposed method outperforms all baseline approaches by increasing the accuracy of scene reconstruction and the grasping success rate.
The advantages of our method include:
1) Unlike simulation-based training, our training-free framework leverages extensive real-world pre-trained knowledge from off-the-shelf models, making it robust in handling a wide variety of objects in real-world scenarios.
2) The framework supports the use of a single-view camera, which can be positioned at non-fixed angles, enhancing its practicality in real-world applications.

In summary, our main contributions are as follows:

\begin{itemize}
    \item We propose a single-view observation pipeline that enables robust depth map reconstruction for transparent and specular objects.
    \item We propose an object mesh replacement mechanism to bridge the gap between single-view object reconstruction techniques and scene depth map reconstruction.
    \item We demonstrate promising performance in both simulation and real-world data, achieving effective results under single-view observation without requiring additional training.
\end{itemize}

\section{RELATED WORK}
\subsection{Transparent and Specular Object Grasping}

Grasping transparent and specular objects presents a significant challenge in robotics~\cite{wang2021graspness,breyer2021volumetric,fang2020graspnet}, primarily due to the incomplete depth information provided by depth cameras.
One common approach to tackle this problem is depth completion or refinement, which aims to improve the noisy and inaccurate depth data before grasp detection~\cite{sajjan2020clear,fang2022transcg,dai2022domain}. 
Other studies~\cite{dai2023graspnerf,kerr2022evo,ummadisingu2024said} take a different approach by eliminating depth sensors altogether, instead leveraging neural radiance field (NeRF) representation to reconstruct transparent shapes from multi-view RGB images for grasping. 
A recent advancement in transparent shape reconstruction~\cite{shi2024asgrasp} further eliminates the need for multiple views, instead utilizing a single RGB image along with raw left-right infrared (IR) pair to reconstruct 3D point cloud for grasping.
Compared to the previous works, our method only requires a single view and is training-free, making it more robust and adaptable to real-world scenarios.

\subsection{Single-view 3D Reconstruction}

The problem of 3D reconstruction from a single view has been a long-standing challenge in the computer vision field. 
Many studies have made significant strides in \textit{scene reconstruction} from single image~\cite{hu2024metric3d,yin2021learning,saxena2008make3d,zhang2023robust}. 
Zhang et al.~\cite{zhang2023robust} introduces a geometry-preserving depth model that provides depth predictions up to an unknown scale. 
Hu et al.~\cite{hu2024metric3d} further develops a foundation model capable of recovering the metric 3D structure of scenes. However, these approaches are typically tailored to everyday objects and often struggle with the reconstruction of transparent and specular objects.
Another line of research explores 3D \textit{object reconstruction} from a single view~\cite{tang2023dreamgaussian,liu2023one,liu2024one,huang2024zeroshape,tochilkin2024triposr,zou2024triplane,he2023openlrm,huang2024zeroshape}.
Liu et al.~\cite{liu2023one,liu2024one} leverages priors of 2D diffusion models to efficiently generate 3D shapes. 
Tang et al.~\cite{tang2023dreamgaussian} adapts 3D Gaussian splatting into generative settings and reduces the image-to-3D optimization time. 
Although these methods can reconstruct transparent and specular objects, most require significant computational time for reconstruction.
In contrast, approaches like TripoSR~\cite{tochilkin2024triposr} accelerate reconstruction, enabling high-quality 3D object generation from a single image within one second, making it a promising candidate for robotic applications.
However, all of these single-view 3D object reconstruction techniques fall short in accurately predicting object pose and scale, which are crucial for integrating the reconstructed object into the original world-coordinate 3D scene for effective robotic manipulation.


\section{METHOD}
\begin{figure*}[htbp]
    \centering
    \includegraphics[width=\textwidth]{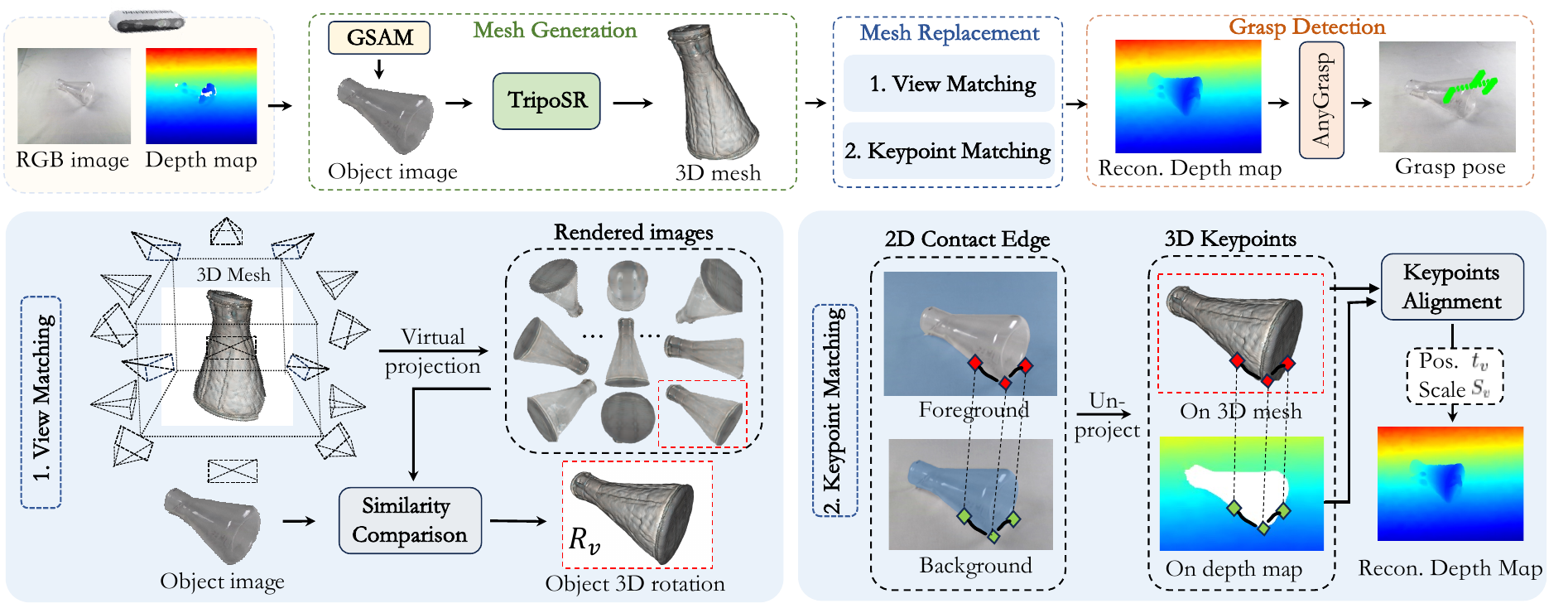}
    \caption{\textbf{The Overall Framework}. The top part illustrates the overall framework flow of reconstructing depth map for transparent or specular objects, while the bottom part details the view matching and keypoint matching modules. These modules work together to determine the \textit{pose} and \textit{scale} required to align the reconstructed 3D object mesh with the captured 3D depth scene.}
    \label{fig:method}
\end{figure*}
\subsection{Problem Formulation}


In this study, we focus on robustly detecting 6-DoF from a single-view RGB-D capture of a real-world scene containing transparent or specular objects, which is challenging due to the lack of clear depth information for such surfaces. 
We first leverage pretrained knowledge from an external single-view 3D object reconstruction approach to obtain an accurate 3D object mesh. 
Our key insight is then to utilize the intrinsic properties of the remaining RGB-D observation, including the depth information from opaque regions and 2D geometry features, to align the 3D mesh with the 3D scene. This aligned mesh subsequently serves as input for the grasp detection module.

Our proposed framework consists of three key modules: \textbf{Mesh Generation} (Sec.~\ref{sec:gen}), \textbf{Mesh Replacement} (Sec.~\ref{sec:replace}), and \textbf{Grasp Detection} (Sec.~\ref{sec:grasp}).
Specifically, given a single RGB-D observation with $I_{rgb} \in \mathbb{R}^{H \times W \times 3}$ and $I_{depth} \in \mathbb{R}^{H \times W}$, the \textbf{Mesh Generation} module uses external models to segment transparent or specular objects $I_{rgb}^{obj.}$ from the captured RGB image $I_{rgb}$ and reconstruct a 3D object mesh $R_{mesh}^{obj.}$. 
The \textbf{Mesh Replacement} module then estimates the 3D object's pose and scale in the scene by generating a rigid transformation matrix $T_v \in SE(3)$ and a scale factor $S_v$ to align the generated 3D mesh $R_{mesh}^{obj.}$ with the corrupted 3D scene $I_{depth}$, resulting in an accurate 3D depth reconstruction for grasping, denoted as $\hat{I}_{depth}$.
The \textbf{Grasp Detection} module calculates optimal 6-DoF grasp poses based on the reconstructed 3D scene $\hat{I}_{depth}$. Each predicted grasp pose is represented as:
\begin{equation}
    g = (t, R, w, q),
\end{equation}
where \( t \in \mathbb{R}^3 \) specifies the 3D position, \( R \in SO(3) \) defines the rotation, $w$ denotes the gripper opening width, and $q$ represents the score for the grasp pose. We then select the grasp pose with the highest score for interaction with the object.

\subsection{Mesh Generation} 
\label{sec:gen}
As illustrated in Fig.~\ref{fig:method}, in this section, we aim to obtain the 3D object reconstruction mesh for transparent or specular objects.
Specifically, given a captured image $I_{rgb}$, we employ Grounded SAM (GSAM)~\cite{liu2024grounding} to segment transparent or specular objects $I_{rgb}^{obj.}$ using text prompts (e.g., ``glass cup").
Next, we apply TripoSR~\cite{tochilkin2024triposr}, which achieves state-of-the-art for fast 3D reconstruction from a single image, even for transparent objects, while maintaining high computational efficiency. Using this method, we reconstruct the 3D object mesh $R_{\text{mesh}}^{\text{obj.}}$ from the segmented object image $I_{\text{rgb}}^{\text{obj.}}$.

\subsection{Mesh Replacement}
\label{sec:replace}
With the geometrically robust 3D object mesh generated, the subsequent challenge is to accurately determine the pose $T_v$ and scale $S_v$ of the 3D mesh in order to place it into the real-world 3D scene for grasp detection. 
Although existing works~\cite{liu2022gen6d,sun2022onepose, he2022onepose++, wen2024foundationpose}, such as Foundation Pose~\cite{wen2024foundationpose}, can predict object pose based on a single-view image and the 3D object mesh, they may fail for transparent or specular objects due to the background pattern noise introduced by these objects. 
Therefore, our goal is to propose a robust module that determines the pose and scale of transparent or specular objects by leveraging accurate depth information from the opaque regions and 2D geometry information of objects. 
Specifically, the mesh replacement module involves two critical steps: \textbf{View Matching} and \textbf{Keypoint Matching}.

\subsubsection{View Matching}. 
This aims to determine the orientation $R_v$ of the 3D mesh such that it aligns with the object orientation captured by the camera. 
Although the 3D information of transparent or specular objects may be inaccurate, the 2D observation reliably reflects the object's 2D shape and geometry.
Therefore, we aim to determine the orientation of the 3D mesh through 2D similarity measurements.

Specifically, we virtually project the 3D mesh $R_{\text{mesh}}^{\text{obj.}}$ onto $N$ views, rendering $N$ images $R_{\text{rgb}} = \{ R_{\text{rgb}}^{1}, \dots, R_{\text{rgb}}^{N} \}$. Next, by calculating the similarity score $S$ between each virtual projected object image $R_{\text{rgb}}^{n}$ and the captured object image $I_{\text{rgb}}^{\text{obj.}}$, we select the virtual projected image with the highest similarity, which reflects the orientation $R_v$ of the 3D mesh in the 3D scene.
The similarity score $S$ consists of three components: \( S_{\text{SSIM}} \), an Structural Similarity Index Measure (SSIM) ~\cite{brunet2011mathematical} that captures the 2D structural similarity of the objects; \( S_{\text{edge}} \), which uses Laplacian edge detection~\cite{torre1986edge} to evaluate 2D edge similarity; and \( S_{\text{ratio}} \), which measures the 2D geometric shape similarity.

For the $S_{\text{SSIM}}$ similarity, we divide the captured object image $I_{\text{rgb}}^{\text{obj.}}$ and the projected object image $R_{\text{rgb}}^{n}$ into $W$ patches and calculate the SSIM similarity~\cite{brunet2011mathematical} for each patch to enhance sensitivity to fine structural details, which are crucial for distinguishing refractive patterns in transparent objects. The equation is as follows:

\begin{equation}
\small
    S_{\text{SSIM}} = \sum_{w \in W} \frac{(2\mu_{w_{I}} \mu_{w_{R}} + C_1)(2\sigma_{w_{I}w_{R}} + C_2)}{(\mu_{w_{I}}^2 + \mu_{w_{R}}^2 + C_1)(\sigma_{w_{I}}^2 + \sigma_{w_{R}}^2 + C_2)}
\end{equation}

, where $\mu_{w_{I}}$ and $\mu_{w_{R}}$ represent the local mean pixel values of the corresponding patches in $I_{\text{rgb}}^{\text{obj.}}$ and $R_{\text{rgb}}^{n}$, respectively. Similarly, $\sigma_{w_{I}}^2$ and $\sigma_{w_{R}}^2$ denote the local variances of the pixel values within the corresponding patches, while $\sigma_{w_{I}w_{R}}$ represents the covariance of the pixel values between the patches. The constants $C_1$ and $C_2$ are used to stabilize the computation.

For the $S_{\text{edge}}$ similarity, we apply Laplacian edge detection $E(\cdot)$~\cite{torre1986edge} to detect the edges of the object in both the captured object image $I_{\text{rgb}}^{\text{obj}}$ and the projected object image $R_{\text{rgb}}^{n}$, and calculate the pixel-wise similarity of the edge images, emphasizing the similarity of geometric contours. The equation is as follows:

\begin{equation}
\small
S_{\text{edge}} = \frac{\sum_{p} E_I(p) \cdot E_R(p)}{\sqrt{\sum_{p} E_I(p)^2 \cdot \sum_{p} E_R(p)^2}}
\end{equation}
, where $E_I$ and $E_R$ represent the edge images of the captured object image $I_{\text{rgb}}^{\text{obj.}}$ and the projected object image $R_{\text{rgb}}^{n}$, respectively, and $p$ denotes each pixel in the image. This formulation ensures robust effectiveness, especially for thin transparent structures, such as bottle necks, where edge contours provide clear geometric features observable in the 2D image.

For $S_{\text{ratio}}$, we aim to measure the similarity in the aspect ratio of the objects:

\begin{equation}
\small
    S_{\text{ratio}} = 1 - \min \left( \frac{|(W_{I}/H_{I}) - (W_{R}/H_{R})|}{(W_{I}/H_{I})}, 1 \right)
\end{equation}
, where $W_I$ and $H_I$ represent the width and height of the object in the captured object image $I_{\text{rgb}}^{\text{obj.}}$, while $W_R$ and $H_R$ represent the width and height of the object in the projected object image $R_{\text{rgb}}^{n}$. This constraint prevents physically implausible matches, such as between tall cylindrical flasks and wide Petri dishes, even if their local textures appear similar.

In total, we select the view $R_{\text{rgb}}^*$ that has the highest similarity score $S$ compared to the captured object image $I_{\text{rgb}}^{\text{obj.}}$, which determines the rotation to obtain view $R_{\text{rgb}}^*$ as the object's rotation $R_{v} \in SO(3)$:

\begin{equation}
\small
\begin{aligned}
    R_{\text{rgb}}^* &= \mathop{\arg\max}\limits_{\{ R_{\text{rgb}}^1, \dots, R_{\text{rgb}}^N \}} S \\
    &= \mathop{\arg\max}\limits_{\{ R_{\text{rgb}}^1, \dots, R_{\text{rgb}}^N \}} \left( \alpha \cdot S_{\text{SSIM}} + \beta \cdot S_{\text{edge}} + \gamma \cdot S_{\text{ratio}} \right)
\end{aligned}
\end{equation}

This view matching process compares over 100 virtually projected views, which are uniformly sampled camera view angles on a spherical surface, thus ensuring a comprehensive evaluation while maintaining high computational efficiency.



\subsubsection{Keypoint Matching}. 
After determining the rotation $R_v$ of the 3D object mesh in the view matching process, we aim to further determine the position $t_v$ and scale $S_v$ of the 3D mesh in the 3D scene, which allows us to place the 3D mesh back into the original 3D scene for robotic grasping.
We observe that while depth information is unreliable for transparent and specular objects, the depth of opaque objects, such as the supporting surface (e.g., a table), remains accurate.
Since the supporting opaque object shares the connecting edge with the transparent or specular objects, the keypoints on the connecting edge from both objects should be at the same 3D position.
Therefore, our goal is to first leverage semantic information from 2D images to identify opaque objects connected to the transparent object. 
We then utilize the contact edge as a 3D positional consistent constraint to match the 3D points from the accurate depth information on the opaque object with the depth of reconstructed 3D mesh, determining the object's position and scale.

Specifically, we identify the 2D contact edge of the transparent or specular objects and their supporting opaque objects (e.g., table) in the captured object image $I_{\text{rgb}}^{\text{obj}}$ by selecting points along the bottom edge of the object. We evenly sample 30 2D keypoints along the contact edge, and then select three keypoints from the left, center, and right regions of these sampled keypoints as the representative 2D keypoints $k_{\text{2D}} = \{k_l, k_m, k_r\}$. As shown in Fig.~\ref{fig:method}, these 2D points exist in both the transparent or specular objects and their supporting opaque objects.

Next, we aim to determine the corresponding 3D points on the opaque objects and the 3D object mesh, which are of positional consistency in 3D if the pose and scale of the 3D object mesh are correct. For the corresponding 3D points on the opaque objects (e.g., table), we leverage the accurate captured depth information of the 2D keypoints $k_{\text{2D}}$ and the camera extrinsics to unproject them into the 3D world coordinates $K_{\text{3D}}^{s}=\{K_l^s, K_m^s, K_r^s\}$. 
Similarly, for the corresponding 3D points on the 3D mesh, we use the depth information corresponding to the 2D keypoints $k_{\text{2D}}$ in the 3D mesh to determine the corresponding 3D points of the object $K_{\text{3D}}^{o}=\{K_l^o, K_m^o, K_r^o\}$. 
Since $K_{\text{3D}}^{s}$ and $K_{\text{3D}}^{o}$ should locate at the same position, and $K_{\text{3D}}^{s}$ on the opaque objects is accurately positioned, we can determine the required translation $t_v$ of the 3D object mesh as follows:
\begin{equation}
t_v = \frac{1}{3} \left( K_{\text{3D}}^{s} - K_{\text{3D}}^{o} \right)
\end{equation}

Subsequently, the overall \(4 \times4\) transformation matrix \(T_v\) to place the 3D object mesh back to the original 3D scene is then constructed as:
\begin{equation}
    T_v = \begin{bmatrix} 
    R_v & t_v \\
    0 & 1 \end{bmatrix}
\end{equation}

As for the scale $S_v$, we use the Euclidean distance $dist(\cdot)$ to compare each corresponding 3D point between the 3D mesh 
 $K_{3D}^{s}$ and the opaque object points $K_{3D}^{o}$:

\begin{equation}
\small
    S_v = \frac{dist(K_{l}^s,K_{m}^s) + dist(K_{l}^s,K_{r}^s) + dist(K_{m}^s,K_{r}^s)}{dist(K_{l}^o,K_{m}^o) + dist(K_{l}^o,K_{r}^o) + dist(K_{m}^o,K_{r}^o)}
\end{equation}

Given $T_v$ and $S_v$, we then obtain an accurate reconstructed 3D depth map $\hat{I}_{depth}$ without transparent or specular corruption.

\subsection{Grasp Detection}
\label{sec:grasp}
We utilize Grasp Anything~\cite{fang2023anygrasp}, which takes the reconstructed 3D depth map $\hat{I}_{depth}$ as input and generates a set of grasp poses. We then filter and select the highest-scoring grasp pose $g$ within the object region for interaction.
While other grasp pose detector such as contact graspnet ~\cite{sundermeyer2021contact} could potentially serve as a drop-in replacement.


\section{Simulation Experiments}
\textbf{Experiment Protocal.}
Following GraspNeRF~\cite{shi2024asgrasp}, we construct the simulation environment using PyBullet~\cite{coumans2016pybullet} for physical grasping simulations and Blender~\cite{fisher2014blender} for photorealistic image rendering. 
It is crucial to note that synthetic depth maps generated from the z-buffer do not accurately capture depth inaccuracies or noise caused by transparent materials. 
To overcome this limitation, we employ a depth sensor simulator~\cite{dai2022domain} to produce simulated depth maps that incorporate realistic sensor noise.
For testing, we use a total of 50 hand-scaled transparent object meshes from~\cite{breyer2021volumetric}, and randomize their textures and materials.

\textbf{Evaluation Metrics.} We evaluate performance using the Success Rate (SR), which is defined as the ratio of successful grasps to the total number of attempts.

\textbf{Baselines Comparisons.}
We compare our method with GraspNeRF~\cite{shi2024asgrasp} in simulation, which is a multi-view, RGB-based 6-DoF grasp detection network that extends the generalizable NeRF model for grasping tasks. 
Our method improves the grasping success rate on transparent objects from 0.75 in the baseline Graspnerf to 0.85, with a 10\% margin.
Notably, unlike the multi-view setting used by the baseline, our approach utilizes only a single-view camera for both reconstruction and grasping.
This shows the effectiveness of our method in generating an accurate reconstruction for robotic grasping.
\begin{figure}[h]
\includegraphics[width=0.49\textwidth]{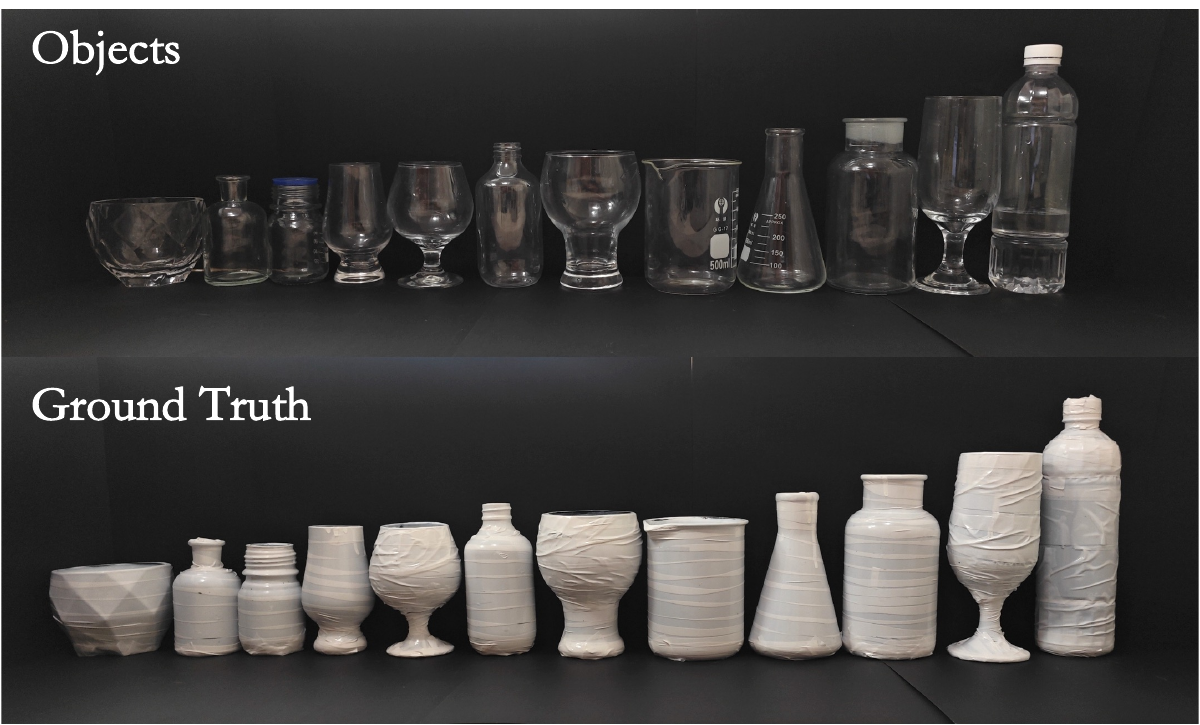}
    \caption{\textbf{Real-world test dataset.} The test dataset consists of 12 transparent objects with diverse shapes. We apply white paint to their surfaces and obtain the ground truth depth maps.}
    \label{fig:real-dataset}
\end{figure}

\section{Real-world Experiments}
\label{sec:real-world}
\subsection{Scene Reconstruction Experiments}
\begin{table}[t]
 \centering
 \caption{Depth reconstruction comparisons on single objects}
\label{tab:recon-single}
\normalsize
 \begin{tabular}{lccc}
\toprule
\rowcolor[HTML]{CBCEFB}
Methods & RMSE$\downarrow$ & REL$\downarrow$ & MAE$\downarrow$ \\
\midrule
Transcg~\cite{fang2022transcg} & 0.1048 & 0.0585 & 0.0715 \\
\rowcolor[HTML]{EFEFEF}
Asgrasp~\cite{shi2024asgrasp} & 0.0918 & 0.0242 & 0.0177   \\
SR3D(Ours) & \textbf{0.0845} & \textbf{0.0156} & \textbf{0.0130}\\
\bottomrule
 \end{tabular}
\end{table}

\begin{table}[h]
 \centering
 \caption{Depth reconstruction comparisons on object in \\cluttered environment}
\label{tab:recon-clutter}
\normalsize
 \begin{tabular}{lccc}
\toprule
\rowcolor[HTML]{CBCEFB}
Methods & RMSE$\downarrow$ & REL$\downarrow$ & MAE$\downarrow$ \\
\midrule
Transcg~\cite{fang2022transcg} & 0.1767 & 0.0885 & 0.0776  \\
\rowcolor[HTML]{EFEFEF}
Asgrasp~\cite{shi2024asgrasp} & 0.1697 & 0.0641 & \textbf{0.0380}   \\
SR3D(Ours) & \textbf{0.1590} & \textbf{0.0564} & 0.0607\\
\bottomrule
 \end{tabular}
\end{table}
\textbf{Evaluation Metrics.} 
Following in Asgrasp~\cite{shi2024asgrasp} and Transcg~\cite{fang2022transcg}, we evaluate the performance of transparent objects depth completion on all objects area by three metrics:
1) RMSE: the root mean squared error,
2) REL: the mean absolute relative difference, 
3) MAE: the mean absolute error. 

\textbf{Test Dataset.}
As shown in Fig.~\ref{fig:real-dataset}, we built a self-collected test dataset consisting of a total of 12 household objects, encompassing various materials shapes. 
For all transparent objects, we apply opaque white spray paint and use a RealSense 435 to capture depth data as ground truth. 
We then compare the reconstruction performance of our method against the baselines. 
Reconstruction evaluations are conducted on both single objects and objects in cluttered environments, as shown in Tab.\ref{tab:recon-single} and Tab.\ref{tab:recon-clutter}, respectively.

\begin{figure*}[t]
    \centering
    \includegraphics[width=\textwidth]{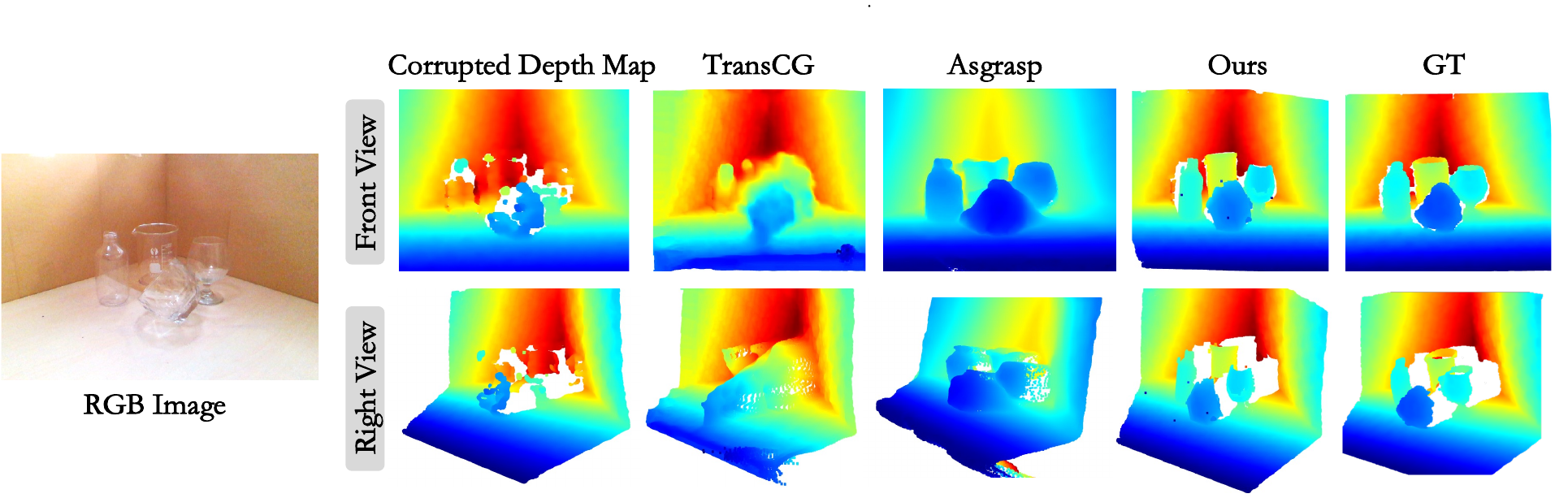}
    \caption{\textbf{Comparisons of depth reconstruction results.} Using the same RGB input, we visualize the reconstruction results of all methods from both the front and right views, providing a clear comparison with the ground truth. }
    \label{fig:recon-viz}
\end{figure*}

\begin{table*}[t]
 \centering
 \caption{Robotic grasping results for transparent objects}
\label{tab:real-grasp}
\normalsize
 \begin{tabular}{lcccccccc}
\toprule
\rowcolor[HTML]{CBCEFB}
 Method & AVG.&Elongated Bottle &Squat Bottle &Conical Flask &Tall Goblet &Phial &Beaker &Short Goblet\\
\midrule
TransCG~\cite{fang2022transcg} &4/10 &5/10 & 3/10 & 4/10 & 5/10 & 4/10  & 2/10 &  2/10\\
\rowcolor[HTML]{EFEFEF}
Asgrasp~\cite{shi2024asgrasp} &6/10 &7/10 & \textbf{8/10} & \textbf{8/10} & 5/10 & 5/10 & 6/10 & 5/10 \\
SR3D (Ours) &\textbf{8/10} &\textbf{9/10} & \textbf{8/10} & 7/10 & \textbf{8/10} & \textbf{9/10} &  \textbf{8/10} & \textbf{8/10}\\
\bottomrule
 \end{tabular}
\end{table*}
\textbf{Result comparisons.}
We compare our method with the following baselines:

1) Transcg~\cite{fang2022transcg}, an end-to-end depth completion network, which takes the RGB image and the inaccurate depth map as inputs and outputs a refined depth map;

2) Asgrasp~\cite{shi2024asgrasp}, an active stereo camera based 6-DoF grasping method for transparent and specular objects. 
It presents a two-layer learning based
stereo network which reconstructs visible and invisible parts of 3D objects. 

As shown in Tab.~\ref{tab:recon-single} and Tab.~\ref{tab:recon-clutter}, comparing with these baseline, 
our method demonstrates promising performance in the real world, as it is training-free and fully leverages the capabilities of off-the-shelf, pretrained single-view object reconstruction models. This allows for robust reconstruction in real-world scenarios. 
Additionally, our mesh replacement mechanism further ensures that the reconstructed object is accurately aligned with the corrupted depth map, resulting in a precise reconstructed scene depth map.
We further visualize the reconstruction result in Fig.~\ref{fig:recon-viz}, where our method is closest to the ground truth.
Although Asgrasp also demonstrates strong 3D geometry reconstruction results, issues such as scale are not well addressed.



\subsection{Grasp Experiments }
\textbf{Real-world Setup.}
We utilize a 7-DoF Franka Panda robotic arm equipped with a 3D printed UMI gripper~\cite{chi2024universal}, along with an Intel RealSense D435 RGB-D camera mounted on the arm. We conduct reconstruction evaluations on both single objects and objects in cluttered environments.
As shown in Tab.~\ref{tab:real-grasp}, we interact with 7 objects.
For each object, we perform 10 grasp trials with varying object poses and camera angles.
We employ the MoveIt!~\cite{gorner2019moveit} motion planner to guide the robotic arm to the target pose. 
A grasp is counted as successful if the object is successfully lifted above the table height and maintained for approximately 5 seconds.

\textbf{Result analysis.}
In Tab.~\ref{tab:real-grasp}, we show the effectiveness of our method in grasping.
Compared to these baselines, our method demonstrates promising performance. 
To ensure a fair comparison and eliminate the possibility that the performance improvement comes from our applied grasp detector, we use the same grasp detector for Asgrasp. 
Specifically, we input the reconstruction results predicted by Asgrasp~\cite{shi2024asgrasp} into AnyGrasp~\cite{fang2023anygrasp} to predict the grasp pose. 
This further highlights the accuracy of our reconstructed depth map, which allows the grasp pose detection model to robustly predict on transparent objects.

\subsection{Ablation Study}
To assess the design of our method, we perform an ablation study to evaluate the effectiveness of the mesh replacement module, including both view matching and keypoint matching mechanisms. We compare it with the Foundation Pose method~\cite{wen2024foundationpose} to highlight the improvements offered by our approach on transparent objects.
The experiments are conducted on single objects in the real-world test dataset, and their performance is evaluated using reconstruction metrics.

In order to assess the effectiveness of our mesh replacement module, we replace it with the Foundation Pose method to determine the object pose in $SE(3)$, which is used to place the generated 3D mesh back into the corrupted 3D scene. For other modules and the acquisition of scale, we maintain consistency with our approach.
As shown in Tab. \ref{tab:recon_abla}, using the foundation pose for object pose estimation leads to a performance drop, compared to our method. 
This performance drop is mainly due to the fact that our experiments are designed for transparent and specular objects, where we leverage reliable depth information from opaque regions along with 2D semantic and geometric cues to determine object 3D pose.
In contrast, the blurriness between the object and background in transparent object images causes confusion for Foundation Pose in determining the object pose, leading to inaccurate pose predictions.


\begin{table}
 \centering
 \caption{Ablation study on depth reconstruction comparisons.}
\label{tab:recon_abla}
\normalsize
 \begin{tabular}{lccc}
\toprule
\rowcolor[HTML]{CBCEFB}
Methods & RMSE$\downarrow$ & REL$\downarrow$ & MAE$\downarrow$ \\
\midrule
FP.~\cite{wen2024foundationpose} & 0.1028 & 0.0211 & 0.0261 \\
SR3D(Ours) & \textbf{0.0823} & \textbf{0.0147} & \textbf{0.0101}\\
\bottomrule
 \end{tabular}
\end{table}



\section{CONCLUSIONS}
We present SR3D, a training-free framework designed for robotic grasping of transparent and specular objects from a single-view observation. 
SR3D leverages external models to generate 3D object reconstructions and accurately localizes objects in the 3D scene using proposed view matching and keypoint matching mechanisms tailored for transparent and specular surfaces. 
Our experiments demonstrate that SR3D outperforms existing methods in both simulated and real-world environments, providing a robust and practical solution for real-world robotic grasping tasks.




\section{Failure Case Analysis and Limitations}
\label{sec:app-fail}
The typical failure modes of the framework can be categorized into two types: a. Low-quality 3D object reconstruction, and b.  Inaccurate pose and scale predictions.

\textit{a. Low-quality 3D object reconstruction.} 
Although TripoSR demonstrates strong performance in most scenarios, it may struggle when the camera view angle is suboptimal, a common challenge in single-view 3D reconstruction. This can undermine the effectiveness of subsequent view matching and keypoint matching modules, which in turn impacts grasp pose prediction. We have observed that TripleSR performs best when the camera view is positioned in the front region. To address this, when inference or manipulation is required, if the camera is not in this optimal range, one possible solution is to reposition the object to a more favorable angle in front of the camera, capture an image for improved mesh reconstruction, and then return the object to its original position for grasping.

\textit{b. Inaccurate pose and scale predictions.}
Most issues arise from the previously mentioned inaccurate mesh reconstruction.
When the reconstructed mesh is flawed, the view matching process will project incorrect images with distorted object shape or geometry, complicating the task of determining the correct object rotation based on 2D geometry similarity comparisons. 
Similarly, poor reconstruction negatively impacts keypoint matching, resulting in misalignment and, as a consequence, errors in scale estimation.

\section{ACKNOWLEDGEMENTS}
This work is supported by The National Youth Talent Support Program (8200800081) and National Natural Science Foundation of China (62376006)

{
\bibliographystyle{IEEEtran}
\bibliography{IEEEabrv,reference}

\begin{thebibliography}{10}
\providecommand{\url}[1]{#1}
\csname url@samestyle\endcsname
\providecommand{\newblock}{\relax}
\providecommand{\bibinfo}[2]{#2}
\providecommand{\BIBentrySTDinterwordspacing}{\spaceskip=0pt\relax}
\providecommand{\BIBentryALTinterwordstretchfactor}{4}
\providecommand{\BIBentryALTinterwordspacing}{\spaceskip=\fontdimen2\font plus
\BIBentryALTinterwordstretchfactor\fontdimen3\font minus \fontdimen4\font\relax}
\providecommand{\BIBforeignlanguage}[2]{{%
\expandafter\ifx\csname l@#1\endcsname\relax
\typeout{** WARNING: IEEEtran.bst: No hyphenation pattern has been}%
\typeout{** loaded for the language `#1'. Using the pattern for}%
\typeout{** the default language instead.}%
\else
\language=\csname l@#1\endcsname
\fi
#2}}
\providecommand{\BIBdecl}{\relax}
\BIBdecl

\bibitem{gervet2023act3d}
T.~Gervet, Z.~Xian, N.~Gkanatsios, and K.~Fragkiadaki, ``Act3d: 3d feature field transformers for multi-task robotic manipulation,'' in \emph{7th Annual Conference on Robot Learning}, 2023.

\bibitem{goyal2023rvt}
A.~Goyal, J.~Xu, Y.~Guo, V.~Blukis, Y.-W. Chao, and D.~Fox, ``Rvt: Robotic view transformer for 3d object manipulation,'' in \emph{Conference on Robot Learning}.\hskip 1em plus 0.5em minus 0.4em\relax PMLR, 2023, pp. 694--710.

\bibitem{goyal2024rvt}
A.~Goyal, V.~Blukis, J.~Xu, Y.~Guo, Y.-W. Chao, and D.~Fox, ``Rvt-2: Learning precise manipulation from few demonstrations,'' \emph{arXiv preprint arXiv:2406.08545}, 2024.

\bibitem{shridhar2023perceiver}
M.~Shridhar, L.~Manuelli, and D.~Fox, ``Perceiver-actor: A multi-task transformer for robotic manipulation,'' in \emph{Conference on Robot Learning}.\hskip 1em plus 0.5em minus 0.4em\relax PMLR, 2023, pp. 785--799.

\bibitem{ze20243d}
Y.~Ze, G.~Zhang, K.~Zhang, C.~Hu, M.~Wang, and H.~Xu, ``3d diffusion policy,'' \emph{arXiv preprint arXiv:2403.03954}, 2024.

\bibitem{grotz2024peract2}
M.~Grotz, M.~Shridhar, Y.-W. Chao, T.~Asfour, and D.~Fox, ``Peract2: Benchmarking and learning for robotic bimanual manipulation tasks,'' in \emph{CoRL 2024 Workshop on Whole-body Control and Bimanual Manipulation: Applications in Humanoids and Beyond}, 2024.

\bibitem{li2024manipllm}
X.~Li, M.~Zhang, Y.~Geng, H.~Geng, Y.~Long, Y.~Shen, R.~Zhang, J.~Liu, and H.~Dong, ``Manipllm: Embodied multimodal large language model for object-centric robotic manipulation,'' in \emph{Proceedings of the IEEE/CVF Conference on Computer Vision and Pattern Recognition}, 2024, pp. 18\,061--18\,070.

\bibitem{liu2024robomamba}
J.~Liu, M.~Liu, Z.~Wang, L.~Lee, K.~Zhou, P.~An, S.~Yang, R.~Zhang, Y.~Guo, and S.~Zhang, ``Robomamba: Multimodal state space model for efficient robot reasoning and manipulation,'' \emph{arXiv preprint arXiv:2406.04339}, 2024.

\bibitem{sundermeyer2021contact}
M.~Sundermeyer, A.~Mousavian, R.~Triebel, and D.~Fox, ``Contact-graspnet: Efficient 6-dof grasp generation in cluttered scenes,'' in \emph{2021 IEEE International Conference on Robotics and Automation (ICRA)}.\hskip 1em plus 0.5em minus 0.4em\relax IEEE, 2021, pp. 13\,438--13\,444.

\bibitem{jia2024lift3d}
Y.~Jia, J.~Liu, S.~Chen, C.~Gu, Z.~Wang, L.~Luo, L.~Lee, P.~Wang, Z.~Wang, R.~Zhang \emph{et~al.}, ``Lift3d foundation policy: Lifting 2d large-scale pretrained models for robust 3d robotic manipulation,'' \emph{arXiv preprint arXiv:2411.18623}, 2024.

\bibitem{keselman2017intel}
L.~Keselman, J.~Iselin~Woodfill, A.~Grunnet-Jepsen, and A.~Bhowmik, ``Intel realsense stereoscopic depth cameras,'' in \emph{Proceedings of the IEEE conference on computer vision and pattern recognition workshops}, 2017, pp. 1--10.

\bibitem{fang2023anygrasp}
H.-S. Fang, C.~Wang, H.~Fang, M.~Gou, J.~Liu, H.~Yan, W.~Liu, Y.~Xie, and C.~Lu, ``Anygrasp: Robust and efficient grasp perception in spatial and temporal domains,'' \emph{IEEE Transactions on Robotics}, vol.~39, no.~5, pp. 3929--3945, 2023.

\bibitem{lu2023vl}
Y.~Lu, Y.~Fan, B.~Deng, F.~Liu, Y.~Li, and S.~Wang, ``Vl-grasp: a 6-dof interactive grasp policy for language-oriented objects in cluttered indoor scenes,'' in \emph{2023 IEEE/RSJ International Conference on Intelligent Robots and Systems (IROS)}.\hskip 1em plus 0.5em minus 0.4em\relax IEEE, 2023, pp. 976--983.

\bibitem{mosbach2024grasp}
M.~Mosbach and S.~Behnke, ``Grasp anything: Combining teacher-augmented policy gradient learning with instance segmentation to grasp arbitrary objects,'' in \emph{2024 IEEE International Conference on Robotics and Automation (ICRA)}.\hskip 1em plus 0.5em minus 0.4em\relax IEEE, 2024, pp. 7515--7521.

\bibitem{curto2022experimental}
E.~Curto and H.~Araujo, ``An experimental assessment of depth estimation in transparent and translucent scenes for intel realsense d415, sr305 and l515,'' \emph{Sensors}, vol.~22, no.~19, p. 7378, 2022.

\bibitem{sajjan2020clear}
S.~Sajjan, M.~Moore, M.~Pan, G.~Nagaraja, J.~Lee, A.~Zeng, and S.~Song, ``Clear grasp: 3d shape estimation of transparent objects for manipulation,'' in \emph{2020 IEEE international conference on robotics and automation (ICRA)}.\hskip 1em plus 0.5em minus 0.4em\relax IEEE, 2020, pp. 3634--3642.

\bibitem{fang2022transcg}
H.~Fang, H.-S. Fang, S.~Xu, and C.~Lu, ``Transcg: A large-scale real-world dataset for transparent object depth completion and a grasping baseline,'' \emph{IEEE Robotics and Automation Letters}, vol.~7, no.~3, pp. 7383--7390, 2022.

\bibitem{dai2022domain}
Q.~Dai, J.~Zhang, Q.~Li, T.~Wu, H.~Dong, Z.~Liu, P.~Tan, and H.~Wang, ``Domain randomization-enhanced depth simulation and restoration for perceiving and grasping specular and transparent objects,'' in \emph{European Conference on Computer Vision}.\hskip 1em plus 0.5em minus 0.4em\relax Springer, 2022, pp. 374--391.

\bibitem{wang2021graspness}
C.~Wang, H.-S. Fang, M.~Gou, H.~Fang, J.~Gao, and C.~Lu, ``Graspness discovery in clutters for fast and accurate grasp detection,'' in \emph{Proceedings of the IEEE/CVF International Conference on Computer Vision}, 2021, pp. 15\,964--15\,973.

\bibitem{breyer2021volumetric}
M.~Breyer, J.~J. Chung, L.~Ott, R.~Siegwart, and J.~Nieto, ``Volumetric grasping network: Real-time 6 dof grasp detection in clutter,'' in \emph{Conference on Robot Learning}.\hskip 1em plus 0.5em minus 0.4em\relax PMLR, 2021, pp. 1602--1611.

\bibitem{fang2020graspnet}
H.-S. Fang, C.~Wang, M.~Gou, and C.~Lu, ``Graspnet-1billion: A large-scale benchmark for general object grasping,'' in \emph{Proceedings of the IEEE/CVF conference on computer vision and pattern recognition}, 2020, pp. 11\,444--11\,453.

\bibitem{dai2023graspnerf}
Q.~Dai, Y.~Zhu, Y.~Geng, C.~Ruan, J.~Zhang, and H.~Wang, ``Graspnerf: Multiview-based 6-dof grasp detection for transparent and specular objects using generalizable nerf,'' in \emph{2023 IEEE International Conference on Robotics and Automation (ICRA)}.\hskip 1em plus 0.5em minus 0.4em\relax IEEE, 2023, pp. 1757--1763.

\bibitem{shi2024asgrasp}
J.~Shi, A.~Yong, Y.~Jin, D.~Li, H.~Niu, Z.~Jin, and H.~Wang, ``Asgrasp: Generalizable transparent object reconstruction and 6-dof grasp detection from rgb-d active stereo camera,'' in \emph{2024 IEEE International Conference on Robotics and Automation (ICRA)}.\hskip 1em plus 0.5em minus 0.4em\relax IEEE, 2024, pp. 5441--5447.

\bibitem{tang2023dreamgaussian}
J.~Tang, J.~Ren, H.~Zhou, Z.~Liu, and G.~Zeng, ``Dreamgaussian: Generative gaussian splatting for efficient 3d content creation,'' \emph{arXiv preprint arXiv:2309.16653}, 2023.

\bibitem{liu2023one}
M.~Liu, C.~Xu, H.~Jin, L.~Chen, M.~Varma~T, Z.~Xu, and H.~Su, ``One-2-3-45: Any single image to 3d mesh in 45 seconds without per-shape optimization,'' \emph{Advances in Neural Information Processing Systems}, vol.~36, pp. 22\,226--22\,246, 2023.

\bibitem{liu2024one}
M.~Liu, R.~Shi, L.~Chen, Z.~Zhang, C.~Xu, X.~Wei, H.~Chen, C.~Zeng, J.~Gu, and H.~Su, ``One-2-3-45++: Fast single image to 3d objects with consistent multi-view generation and 3d diffusion,'' in \emph{Proceedings of the IEEE/CVF conference on computer vision and pattern recognition}, 2024, pp. 10\,072--10\,083.

\bibitem{huang2024zeroshape}
Z.~Huang, S.~Stojanov, A.~Thai, V.~Jampani, and J.~M. Rehg, ``Zeroshape: Regression-based zero-shot shape reconstruction,'' in \emph{Proceedings of the IEEE/CVF Conference on Computer Vision and Pattern Recognition}, 2024, pp. 10\,061--10\,071.

\bibitem{tochilkin2024triposr}
D.~Tochilkin, D.~Pankratz, Z.~Liu, Z.~Huang, A.~Letts, Y.~Li, D.~Liang, C.~Laforte, V.~Jampani, and Y.-P. Cao, ``Triposr: Fast 3d object reconstruction from a single image,'' \emph{arXiv preprint arXiv:2403.02151}, 2024.

\bibitem{zou2024triplane}
Z.-X. Zou, Z.~Yu, Y.-C. Guo, Y.~Li, D.~Liang, Y.-P. Cao, and S.-H. Zhang, ``Triplane meets gaussian splatting: Fast and generalizable single-view 3d reconstruction with transformers,'' in \emph{Proceedings of the IEEE/CVF conference on computer vision and pattern recognition}, 2024, pp. 10\,324--10\,335.

\bibitem{he2023openlrm}
Z.~He and T.~Wang, ``Openlrm: Open-source large reconstruction models,'' 2023.

\bibitem{liu2024grounding}
S.~Liu, Z.~Zeng, T.~Ren, F.~Li, H.~Zhang, J.~Yang, Q.~Jiang, C.~Li, J.~Yang, H.~Su \emph{et~al.}, ``Grounding dino: Marrying dino with grounded pre-training for open-set object detection,'' in \emph{European Conference on Computer Vision}.\hskip 1em plus 0.5em minus 0.4em\relax Springer, 2024, pp. 38--55.

\bibitem{kerr2022evo}
J.~Kerr, L.~Fu, H.~Huang, Y.~Avigal, M.~Tancik, J.~Ichnowski, A.~Kanazawa, and K.~Goldberg, ``Evo-nerf: Evolving nerf for sequential robot grasping of transparent objects,'' in \emph{6th annual conference on robot learning}, 2022.

\bibitem{ummadisingu2024said}
A.~Ummadisingu, J.~Choi, K.~Yamane, S.~Masuda, N.~Fukaya, and K.~Takahashi, ``Said-nerf: Segmentation-aided nerf for depth completion of transparent objects,'' \emph{arXiv preprint arXiv:2403.19607}, 2024.

\bibitem{hu2024metric3d}
M.~Hu, W.~Yin, C.~Zhang, Z.~Cai, X.~Long, H.~Chen, K.~Wang, G.~Yu, C.~Shen, and S.~Shen, ``Metric3d v2: A versatile monocular geometric foundation model for zero-shot metric depth and surface normal estimation,'' \emph{IEEE Transactions on Pattern Analysis and Machine Intelligence}, 2024.

\bibitem{yin2021learning}
W.~Yin, J.~Zhang, O.~Wang, S.~Niklaus, L.~Mai, S.~Chen, and C.~Shen, ``Learning to recover 3d scene shape from a single image,'' in \emph{Proceedings of the IEEE/CVF Conference on Computer Vision and Pattern Recognition}, 2021, pp. 204--213.

\bibitem{saxena2008make3d}
A.~Saxena, M.~Sun, and A.~Y. Ng, ``Make3d: Learning 3d scene structure from a single still image,'' \emph{IEEE transactions on pattern analysis and machine intelligence}, vol.~31, no.~5, pp. 824--840, 2008.

\bibitem{zhang2023robust}
C.~Zhang, W.~Yin, G.~Yu, Z.~Wang, T.~Chen, B.~Fu, J.~T. Zhou, and C.~Shen, ``Robust geometry-preserving depth estimation using differentiable rendering,'' in \emph{Proceedings of the IEEE/CVF International Conference on Computer Vision}, 2023, pp. 8951--8961.

\bibitem{liu2022gen6d}
Y.~Liu, Y.~Wen, S.~Peng, C.~Lin, X.~Long, T.~Komura, and W.~Wang, ``Gen6d: Generalizable model-free 6-dof object pose estimation from rgb images,'' in \emph{European Conference on Computer Vision}.\hskip 1em plus 0.5em minus 0.4em\relax Springer, 2022, pp. 298--315.

\bibitem{sun2022onepose}
J.~Sun, Z.~Wang, S.~Zhang, X.~He, H.~Zhao, G.~Zhang, and X.~Zhou, ``Onepose: One-shot object pose estimation without cad models,'' in \emph{Proceedings of the IEEE/CVF Conference on Computer Vision and Pattern Recognition}, 2022, pp. 6825--6834.

\bibitem{he2022onepose++}
X.~He, J.~Sun, Y.~Wang, D.~Huang, H.~Bao, and X.~Zhou, ``Onepose++: Keypoint-free one-shot object pose estimation without cad models,'' \emph{Advances in Neural Information Processing Systems}, vol.~35, pp. 35\,103--35\,115, 2022.

\bibitem{wen2024foundationpose}
B.~Wen, W.~Yang, J.~Kautz, and S.~Birchfield, ``Foundationpose: Unified 6d pose estimation and tracking of novel objects,'' in \emph{Proceedings of the IEEE/CVF Conference on Computer Vision and Pattern Recognition}, 2024, pp. 17\,868--17\,879.

\bibitem{brunet2011mathematical}
D.~Brunet, E.~R. Vrscay, and Z.~Wang, ``On the mathematical properties of the structural similarity index,'' \emph{IEEE Transactions on Image Processing}, vol.~21, no.~4, pp. 1488--1499, 2011.

\bibitem{torre1986edge}
V.~Torre and T.~A. Poggio, ``On edge detection,'' \emph{IEEE Transactions on Pattern Analysis and Machine Intelligence}, no.~2, pp. 147--163, 1986.

\bibitem{coumans2016pybullet}
E.~Coumans and Y.~Bai, ``Pybullet, a python module for physics simulation for games, robotics and machine learning,'' 2016.

\bibitem{fisher2014blender}
G.~Fisher, \emph{Blender 3D Basics Beginner's Guide: A quick and easy-to-use guide to create 3D modeling and animation using Blender 2.7}.\hskip 1em plus 0.5em minus 0.4em\relax Packt Publishing Ltd, 2014.

\bibitem{chi2024universal}
C.~Chi, Z.~Xu, C.~Pan, E.~Cousineau, B.~Burchfiel, S.~Feng, R.~Tedrake, and S.~Song, ``Universal manipulation interface: In-the-wild robot teaching without in-the-wild robots,'' \emph{arXiv preprint arXiv:2402.10329}, 2024.

\bibitem{gorner2019moveit}
M.~G{\"o}rner, R.~Haschke, H.~Ritter, and J.~Zhang, ``Moveit! task constructor for task-level motion planning,'' in \emph{2019 International Conference on Robotics and Automation (ICRA)}.\hskip 1em plus 0.5em minus 0.4em\relax IEEE, 2019, pp. 190--196.

\end{thebibliography}
}

\end{document}